\documentclass[10pt,twocolumn,letterpaper]{article}

\usepackage{iccv}
\usepackage{times}
\usepackage{epsfig}
\usepackage{graphicx}
\usepackage{amsmath}
\usepackage{amssymb}
\usepackage{outlines}
\usepackage{booktabs}
\usepackage{nicematrix}
\usepackage{multirow}
\usepackage[pagebackref=true,breaklinks=true,letterpaper=true,colorlinks,bookmarks=false]{hyperref}

\iccvfinalcopy % *** Uncomment this line for the final submission

\def\iccvPaperID{3} % *** Enter the ICCV Paper ID here

\ificcvfinal\fi

\newcommand{\methodname}{SEPAL}
\usepackage{comment}

\begin{document}

%%%%%%%%% TITLE
\title{\methodname: Spatial Gene Expression Prediction from Local Graphs}

\author{Gabriel Mejia, Paula Cárdenas$^*$, Daniela Ruiz$^*$, Angela Castillo, Pablo Arbeláez\\
Center for Research and Formation in Artificial Intelligence\\
Universidad de los Andes, Bogotá, Colombia\\
{\tt\small \{gm.mejia,p.cardenasg,da.ruizl1,a.castillo13,pa.arbelaez\}@uniandes.edu.co}
% \and
}

\maketitle
\def\thefootnote{*}\footnotetext{These authors contributed equally to this work}\def\thefootnote{\arabic{footnote}}
\ificcvfinal\thispagestyle{empty}\fi

%%%%%%%%% ABSTRACT
\begin{abstract}
   Spatial transcriptomics is an emerging technology that aligns histopathology images with spatially resolved gene expression profiling. It holds the potential for understanding many diseases but faces significant bottlenecks such as specialized equipment and domain expertise. In this work, we present \methodname, a new model for predicting genetic profiles from visual tissue appearance. Our method exploits the biological biases of the problem by directly supervising relative differences with respect to mean expression, and leverages local visual context at every coordinate to make predictions using a graph neural network. This approach closes the gap between complete locality and complete globality in current methods. In addition, we propose a novel benchmark that aims to better define the task by following current best practices in transcriptomics and restricting the prediction variables to only those with clear spatial patterns. Our extensive evaluation in two different human breast cancer datasets indicates that \methodname~outperforms previous state-of-the-art methods and other mechanisms of including spatial context.
\end{abstract}

%%%%%%%%% BODY TEXT
\section{Introduction}
\begin{comment}
\1 H\&E imaging and gene expression introduction:
        \2 Eosin and Ematoxin histopathological images are currently the gold standards for the majority of diseases
        \2 In the last 2 decades this type of information has been complemented with additional quantification of molecular biomarkers such as mRNA expression levels
        \2 The expression levels have the advantage of being highly specific for diseases and can directly influence prognosis and treatment
        \2 Both domains are complementary because H\&E imaging lacks the specificity of gene quantification whereas gene profiling inherently lacks the functional and physiological insights derived from spatial morphology observed in images.
        \2 Recognising that both gene expression and morphology are phenomenons deeply influenced by spatial information, a new set of technologies, collectively known as spatial transcriptomics (ST), have recently emerged that align dense spatially resolved transcriptomic profiling with H\&E images.
\end{comment}
Histopathology is the study of diseases in tissues through microscopic sample examination. Among the different staining methods, Hematoxylin and Eosin (H\&E) is the most common one and is currently considered the gold standard for diagnosing a wide range of diseases~\cite{gold_standard1,gold_standard2,gold_standard3}. More recently, this approach has been complemented with molecular biomarkers, such as mRNA expression profiling, offering high specificity and the ability to directly predict prognosis and determine treatments~\cite{prognosis1,prognosis2}. Interestingly, these two data types prove complementary: while H\&E imaging lacks the specificity of transcriptomics, gene profiling lacks the physiological insights derived from morphology. By aligning dense spatial mRNA profiling with H\&E histopathological images, Spatial Transcriptomics technologies (ST) provide comprehensive insights into the spatial organization of gene expression within tissues~\cite{ST_technologies}.

\begin{comment}
    \1 Spatial transcriptomics potential and problems
        \2 Now that gene expression can be checked directly on the tissue morphology, these novel datasets hold the potential for an unprecedented understanding of the mechanistic causes behind many diseases.
        \2 However, as with any emerging technology, the specialized equipment, domain expertise, and time required to perform these techniques adequately are major bottlenecks for this technology to benefit patients in real clinical practice.
\end{comment}

The advent of direct gene expression assessment on tissue harbors the potential for an unprecedented understanding of the mechanistic causes behind many diseases. 
However, obtaining these datasets in real clinical practice encounters major bottlenecks, primarily stemming from the need for specialized equipment, domain expertise, and considerable time requirements~\cite{challenges}. To overcome these burdens and leverage the fact that H\&E images are ubiquitous in medical settings, the computer vision community has recently delved into predicting gene expression from tissue images. Although various works demonstrate promising results~\cite{HE2RNA,HisToGene,tRNAsformer,STNet,EGN,EGGN}, existing methods are still far from clinical deployment. 

\begin{comment}
    \1 Task formulation and short introduction to methods
        \2 To overcome the burdens of these methods, and leverage the fact that H\&E images are ubiquitous in medical diagnosis, the computer vision community has been recently studying the task of predicting gene expression from histopathological images.
        \2 Various works have been developed with promising results (cite cite cite)
        \2 Although the potential is clear, methods are still far behind the performance levels required for clinical practice
        \2 The causes of that can be divided into inherent problem challenges and approach (methods) limitations.
\end{comment}

\begin{figure*}
    \centering
    \includegraphics[width=0.99\textwidth]{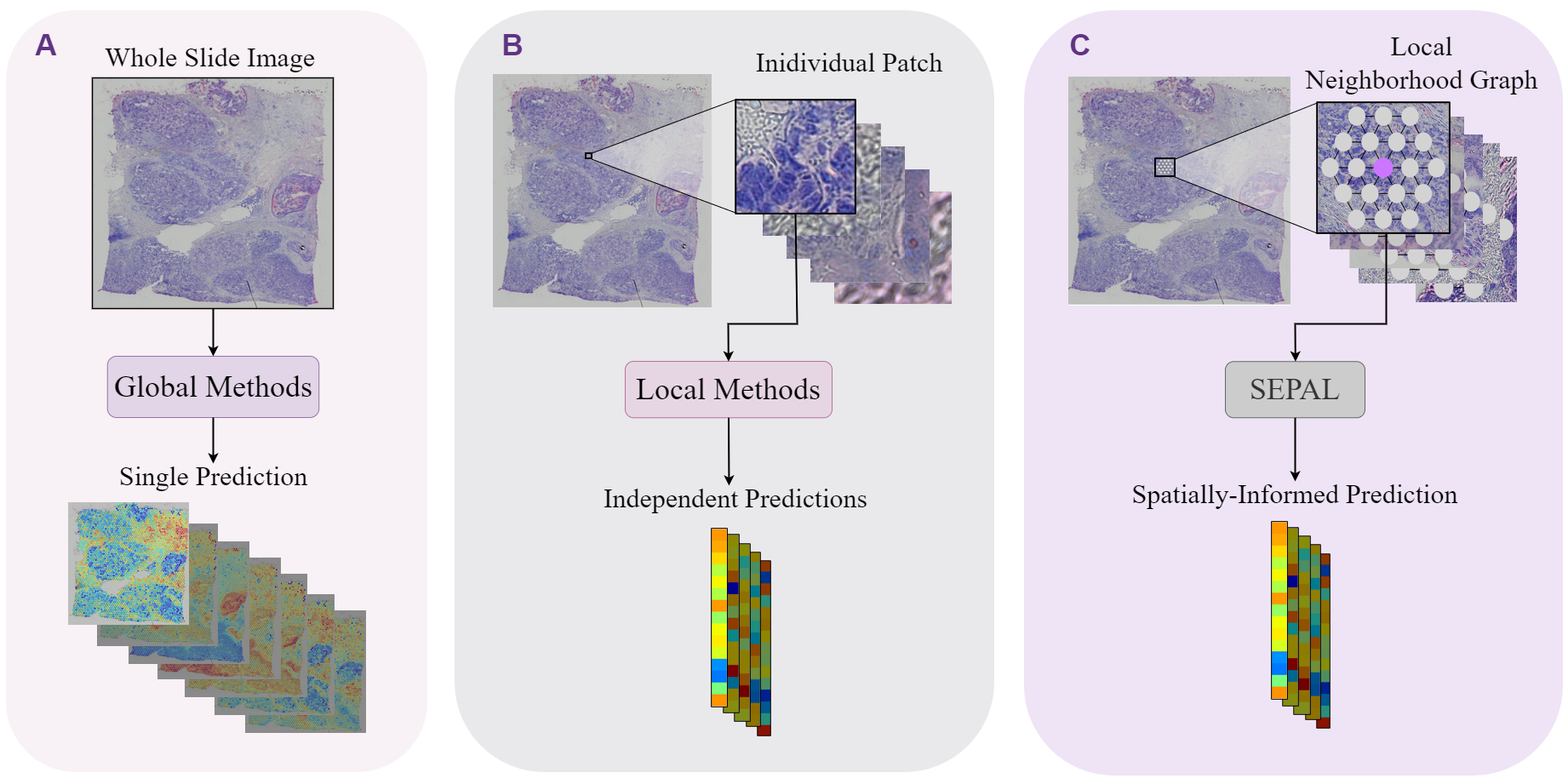}
    \caption{Different approaches for predicting gene expression from tissue images. The inputs of each model type are enclosed by a black frame. (A) \emph{Global methods} analyze a whole slide image and make a prediction about the tissue expression in every spot at once. (B) \emph{Local methods} process the image by patches and predict the expression of each individual patch, one at a time. (C) \methodname~uses graphs that contain information from multiple patches to represent spatial information and predict gene expression for the central node of each graph.}
    \label{fig:overview}
\end{figure*}

\begin{comment}
    \1 Observed problems in the task itself
        \2 A dependence between the appearance of tissue and the expression of a given gene is unnecessary, and there could be nothing to predict (\eg constitutive genes). [Constitutive genes being independent of visual appearance example]. In other words, if a gene has an expression pattern that does not change with the spatial context, it might not be suitable for prediction based on visual information.
        \2 The current publicly available datasets contain $2-70$ whole slide images with $5,000-15,000$ gene expression values for a set of $300-3500$ coordinates depending on the technology.
        \2 Generating such high-dimensional predictions with very few samples is inherently difficult.
        \2 Groundtruth data is sparse and noisy because the technology is still in development
\end{comment}

Upon closer examination of the problem, it becomes evident that changes in gene expression are typically associated with alterations in tissue appearance. However, it is important to note that this correlation does not necessarily apply to all genes. For example, constitutive genes that exhibit constant expression within the spatial context~\cite{constitutive_genes} are unsuitable for prediction based solely on visual information. Hence, methods should focus on genes with a verifiable dependence on tissue appearance for the task to be well defined. Another challenge lies in the scarcity of data.  The current publicly available datasets encompass $2-70$ Whole Slide Images (WSI) with $5,000-15,000$ genes for a set of $300-3500$ coordinates, depending on the technology~\cite{ST_technologies}. Consequently, generating such high-dimensional predictions with such limited samples is intrinsically difficult.  Finally, as the technology is still in development, ground-truth data is sparse and noisy; specifically, pepper noise is observed in the expression maps \cite{challenges}.

\begin{comment}
    \1 Observed problems in the current methods
        \2 State of the art usually focuses on the prediction of the $\approx250$ most expressed genes. However, as noted previously, a constant, highly expressed gene will pass this filtering without being a suitable prediction variable.
        \2 Dichotomy between complete locality and complete globality.
        \begin{itemize}
            \item Complete globality leverages spatial information. However, it faces data scarcity, which makes models more prone to overfitting. 
            \item Complete locality has all data needed for deep learning training, but overlooks the spatial relations, which leads to suboptimal performance.
        \end{itemize}
\end{comment}

Current approaches present a dichotomy between complete globality, which uses the WSI to jointly predict an expression map for all the coordinates at once (WSI-based methods \cite{HE2RNA, HisToGene, tRNAsformer}, Fig.\ref{fig:overview}.A), and complete locality, which only uses visual information available at each coordinate to predict gene expression (patch-based methods \cite{STNet, EGN, EGGN}, Fig.\ref{fig:overview}.B). 
While complete globality leverages spatial information and long-range interactions, it suffers from severe data scarcity, making models prone to overfitting. In contrast, complete locality benefits from abundant data for deep learning training but disregards spatial relations, resulting in suboptimal performance. % Besides that, current works focus on the prediction of the $\approx250$ most expressed genes. And, as noted previously, a constant, highly expressed gene will pass that filtering without being suitable for prediction.

\begin{comment}
    \1 Our proposed method
        \2 Addressing all the identified challenges, we propose a new problem formulation, benchmark, and state-of-the-art method: \methodname (standing for XXXX).
        \2 Our problem formulation allows us to take advantage of the biological nature of the problem. Our benchmark uses a strong bioinformatic pipeline and classical vision ideas to overcome the problems of gene predictability and noisy ground truths. And our model closes the gap between locality and globality by performing local spatial analysis.
\end{comment}

To overcome these challenges, we propose a new problem formulation, benchmark, and state-of-the-art method for \textbf{S}patial \textbf{E}xpression \textbf{P}rediction by \textbf{A}nalysing \textbf{L}ocal graphs (\textbf{\methodname}). Our problem formulation  strategically exploits the biological nature of the problem; our benchmark uses a robust bioinformatic pipeline to overcome acquisition issues; and our model bridges the gap between locality and globality by performing local spatial analysis.

\begin{comment}
    \1 Reasoning behind delta prediction
        \2 Biologically, the expression of a gene is expected to be defined within a specific range of values, and the variations inside that range are the ones with physiological significance rather than the absolute value of the expression.
        \2 In practice, this means that the prediction space is bounded within a box whose center can be used as an inductive bias.  
        \2 Given that this bias is easily estimated from training data, we focus on learning variations from those values based on visual information. This differs from previous works since they aim to directly learn the absolute expression. 
        \2 Namely, we focus on predicting delta variations from the mean expression of each gene in the training dataset.    
\end{comment}

In terms of problem formulation, we leverage a domain-specific advantage: the expression of a gene is expected to be within a specific range of values, and the variations inside that range are the ones with physiological significance. Rather than solely focusing on the absolute value of gene expression, we exploit this knowledge by bounding the prediction space within a defined box and using its center as an inductive bias. By estimating this bias from the training data, we can focus on learning relative differences instead of absolute values. Specifically, we supervise expression changes with respect to the mean expression of each gene in the training dataset. This novel approach differs from previous works since they directly predict the absolute gene expression.

\begin{comment}
    \1 Why we overcome the challenges benchmark
        \2 For the benchmark we incorporate standard bioinformatic processing which was lacking previously (possibly due to little interaction between the areas)
        \2 We apply a modified version of an adaptive median filter to deal with the pepper noise
        \2 We focused on predicting the genes that present significant spatial variation assessed by Moran's I Statistic.
\end{comment}

We build our benchmark by first incorporating standard bioinformatic processing normalizations (TPM \cite{TPMs}), which were previously lacking. Then, we apply a modified version of an adaptive median filter \cite{adaptive_filter} to manage the pepper noise. And finally, to ensure the selection of relevant prediction genes, we filter by Moran's I \cite{moran} value, a statistic designed to identify significant spatial patterns over a graph. By leveraging Moran's I, we ensure our focus remains on genes that depend on tissue appearance.

\begin{comment}
    \1 Why we overcome the challenges method
        \2 We performed local spatial analysis using graph neural networks in k-hop neighborhoods of each patch.
        \2 With this approach, we have the advantages of having enough training data and being able to use spatial interactions
        \2 We used 2 datasets with different technologies and spatial scales of human breast cancer
        \2 We obtained state state-of-the-art performance
\end{comment}

Lastly, we introduce a novel approach that harnesses the power of local spatial analysis. Our strategy starts with a completely local learning stage and then integrates information from local neighborhoods surrounding each patch with the help of a graph neural network (GNN, see Fig.\ref{fig:overview}.C). Our key hypothesis is that gene expression is predominantly influenced by nearby visual characteristics rather than long-range interactions. \methodname~benefits from the advantages of local-based and global-based training (spatial relations and enough data) without succumbing to their respective limitations. We conduct extensive experimentation on two different human breast cancer datasets obtained with different technologies and report favorable results relative to existing techniques.

%\cite{STNet_dataset1,STNet_dataset2}

Our contributions can be summarized as follows:
\begin{itemize}
    \item We propose a paradigm shift to supervise gene expression changes relative to the mean rather than absolute values.
    \item We propose a benchmark that follows current best practices in transcriptomics, deals with pepper noise, and restricts prediction genes to only those with clear spatial patterns.
    \item We develop a new state-of-the-art method that applies local spatial analysis via graph neural networks.
\end{itemize}

To promote further research on ST, our project's benchmark and source code is publicly available at \url{https://github.com/BCV-Uniandes/SEPAL}.

%----------------------------------------------------------------------------------------------------------
\section{Related Work}
Multiple approaches have been proposed to tackle the gene expression prediction task, with works focusing on different aspects of the visual data. State-of-the-art methods can be divided into two paradigms: global (WSI-based) and local (patch-based) focused.

\subsection{Global Methods}
Global methods predict the gene expression of all the spots of a WSI at once, meaning that their input corresponds to the complete data from a high-resolution histopathology image. The most notable work of this family of methods is HisToGene \cite{HisToGene}, which receives a WSI and divides it into patches that are represented through image and positional embeddings fed to a Vision Transformer (ViT) architecture \cite{ViT}. 

The mechanism in HisToGene enables the model to consider spatial associations between spots \cite{HisToGene}. Nonetheless, this method demands many WSIs, posing a challenge as WSIs are often scarce in most datasets. Additionally, processing the entire WSI incurs in a high computational cost. Therefore, we propose a more efficient spatial analysis at a smaller scale. Instead of using an entire sample as a single data element, we adopt a patch-based strategy, enabling us to execute predictions one patch at a time. This granular approach not only conserves computational resources but also mitigates the overfitting risks associated with using large WSIs.  

\subsection{Local Methods} 
Unlike global methods, local methods estimate the gene expression one spot at a time by dividing the WSI into individual patches. Some examples of this approach include STNet \cite{STNet}, EGN \cite{EGN}, and EGGN \cite{EGGN}. The focal point of local methods is the visual information in the patch of interest, and they do not take into consideration characteristics such as the vicinity of the patch or long-range interactions.

For instance, STNet \cite{STNet}, which is one of the most popular methods, formulates the task as a multivariate regression problem, and its architecture consists of a finetuned CNN (DenseNet-121 \cite{DenseNet}) whose final layer is replaced with a linear layer that predicts the expression of 250 genes. A characteristic strategy of STNet is that during inference, it predicts the gene expression for 8 different symmetries of that image (4 rotation angles and their respective reflections) and returns the mean result as the final estimation. This model generalizes well across datasets and has high performance when predicting the spatial variation in the expression of well-known cancer biomarkers \cite{STNet}.

Other examples of this approach include EGN \cite{EGN} and its upgraded version, EGGN \cite{EGGN}. The core of these methods is exemplar guidance learning \cite{EGN}, a tool that they apply to base their predictions on the expressions of the patches that are most visually similar to the patch of interest. These reference patches are known as the exemplars and correspond to the nearest neighbors of a given patch in the latent space of an image encoder. The difference between these two models lies in the main processing of the input, where EGN uses the exemplars to guide a ViT \cite{ViT}, while EGGN uses the exemplars to build visual similarity graphs that are fed to a GraphSAGE-based backbone \cite{GraphSAGE}.

The key hypothesis of EGN and EGGN is that similar images have similar gene expression patterns, no matter their location within a tissue. Nevertheless, depending on the scale of the patches, this assumption could neglect their local context. For instance, if each patch contains a single cell, several similar patches with different physiological contexts might differ in their transcriptomic profile. Thus, for our approach we choose to guide our model with spatially close patches rather than with visually similar patches. When we consider the surroundings of a specific patch, we take into account its location within the tissue and the possible differences in its biological profile. As a result, we tackle the potential limitation that the scale of the input could impose.

\section{\methodname}

\subsection{Problem Formulation}

Given an input image patch $X \in \mathbb{R}^{[H,W,3]}$, and $k$ spatial neighbors $Z \in \mathbb{R}^{[k,H,W,3]}$, we want to train an estimator $F_{\theta}(\cdot)$ that predicts the difference between the gene expression $y$ of patch $X$ and the mean expressions in the training set $\bar{y}_{\text{train}}$. Consequently, we aim to optimize a set of parameters $\theta^*$ such that:
\begin{align}
    F_{\theta^*}(X, Z) \approx \Delta y = y - \bar{y}_{\text{train}} 
\end{align}

Where, $\Delta y \in \mathbb{R}^{[n_g, 1]}$ is the difference between $\bar{y}_{\text{train}} \in \mathbb{R}^{[n_g, 1]}$ and the real gene expression $y \in \mathbb{R}^{[n_g, 1]}$ of the patch. This paradigm shift of predicting $\Delta y$ instead of $y$, has the purpose of allowing our method to focus directly on the nuances in the data since we are centering the dynamic range of the prediction space around zero.

% TODO: F se divide en I (image encoder): [H,W,3] -> emd_dim, y que luego hay una capa lineal
% S (spatial) [N, emd_dim] -> n_genes 
% Donde primero se preentrena I para obtener predicciones iniciales a nivel completamente local y luego se utiliza S para refinar estas predicciones y obtener una correccion que tome encuenta la informacion espacial local

\subsection{Architecture Overview}
\label{sec:architecture}

\begin{figure*}
    \centering
    \includegraphics[width=0.97\textwidth]{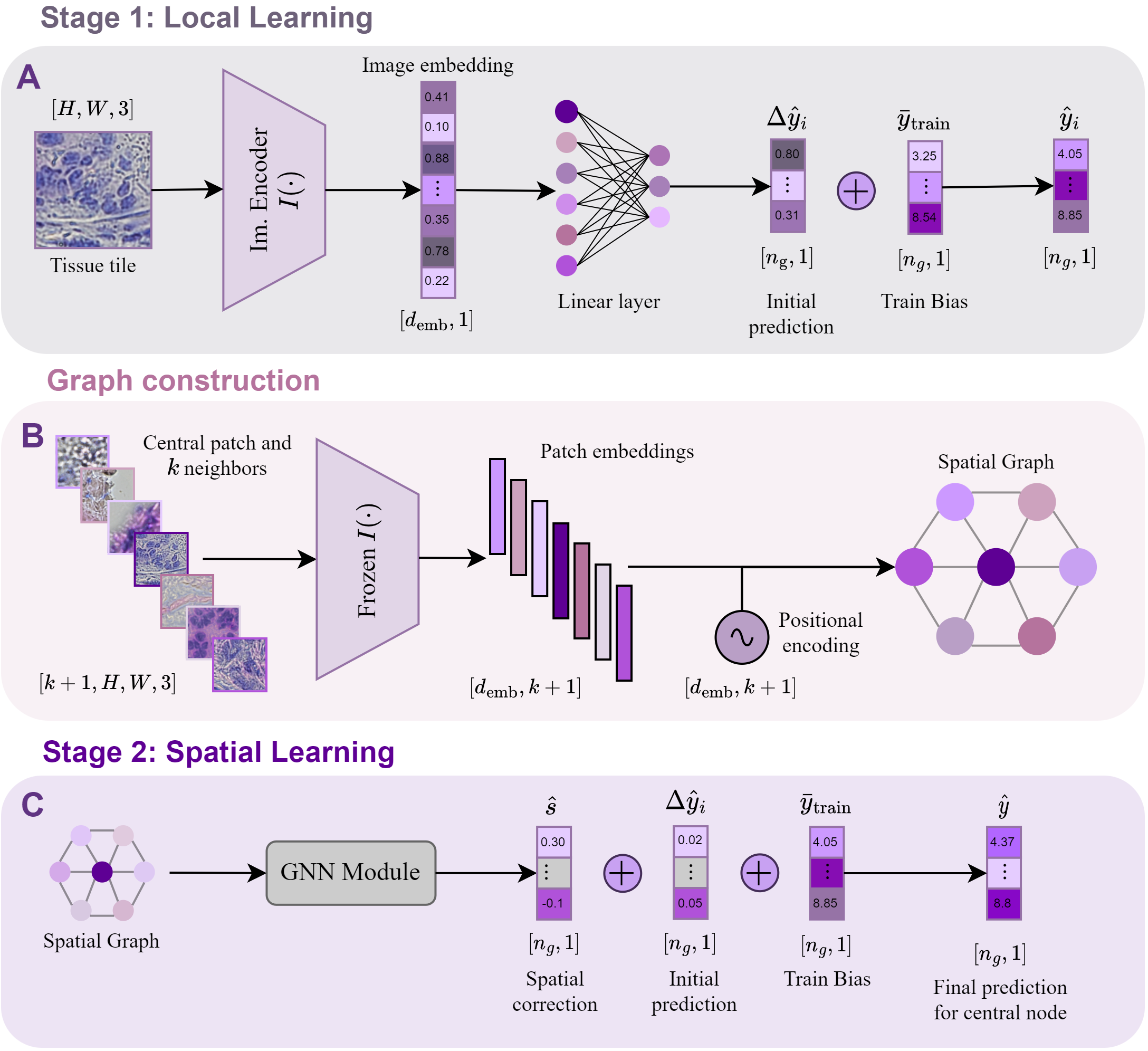}
    \caption{(A)First stage of our proposal. Pretraining of the Image Encoder $I(\cdot)$ and a linear layer $L(\cdot)$ to output the Image Embedding $(I_{\text{emb}})$ of a patch $X$, along with a preliminar prediction $\Delta \hat{y}_{i}$ of the difference between the expression in the patch and the mean expression in the train dataset.  (B) The Graph Construction process begins with an image patch of interest and its spatial neighbors to build the graph representation based on the patch embeddings returned by the frozen $I(\cdot)$ and the positional encoding of each neighbor. (C) Architecture of the spatial learning module, which receives as input a Spatial Graph of the patch neighborhood and applies a GNN to predict the spatial correction $\hat{s}$ that further improves the $\Delta \hat{y}_{i}$ to get the $\Delta \hat{y}$ associated to the center patch of the graph and obtain the final gene expression prediction $\hat{y}$.}
    \label{fig:model_overview}
\end{figure*}

\methodname~is comprised of two stages: local learning and spatial learning, which are shown in Fig.\ref{fig:model_overview}. While local learning follows the classic approach of finetuning and image encoder, the spatial learning of \methodname~relies on representing the input patch and its neighbors as a graph, where the central node corresponds to the image for which we want to predict the gene expression. With this representation, our model has access to the visual features in the current location and in its surroundings.

Prior to the construction of the graphs, in the first stage of our proposal (Fig.\ref{fig:model_overview}.A), we train a feature extractor $I(\cdot)$ to process an input image patch $X$ and return a low-dimensional embedding $I_{\text{emb}} \in \mathbb{R}^{\left[ d_{\text{emb}}, 1\right]}$. Besides, this module also outputs a local prediction $\Delta \hat{y}_{i} \in \mathbb{R}^{[n_g, 1]}$ obtained by applying a linear layer $L(\cdot)$ to $I_{\text{emb}}$ as follows: 
\begin{align}
    &I(X) = I_{\text{emb}}\\
    &L(I_{\text{emb}}) = \Delta \hat{y}_{i} \approx y - \bar{y}_{\text{train}}
\end{align}

Consequently, the preliminary prediction $\Delta \hat{y}_{i}$ is completely based on $X$ and is later refined in the spatial learning stage. After training $I(\cdot)$, we fix it and use it to obtain the visual features of all the patches in the dataset. We integrate these embeddings, together with a transformer-like positional encoding, to construct a local neighborhood graph $\mathcal{G}(X)$ for each patch (Fig.\ref{fig:model_overview}.B).

%for graph construction (Details in Section \ref{sec:graph_construction}). 
% (Figure \ref{fig:model_overview}C) the model receives as input one graph per patch $\mathcal{G}(V, E, M)$ that contains the set $V$ of $k+1$ nodes, the binary set of undirected edges $E$, and the node feature matrix $M \in \mathbb{R}^{[d_{\text{emb}}, k+1]}$. The 

Lastly, in the spatial learning stage (Fig.\ref{fig:model_overview}.C), input graphs are processed by a GNN Module to obtain a spatial correction vector $\hat{s} \in \mathbb{R}^{\left[ n_{g}, 1\right]}$ which is then added to $\Delta \hat{y}_{i}$ to obtain $\Delta \hat{y}$. This spatially aware prediction is summed with the bias $\bar{y}_{\text{train}}$ to present the final gene expression estimation $\hat{y}$ for the input patch:
\begin{align}
    &\Delta \hat{y} = \hat{s} + \Delta \hat{y}_{i}\\
    &\hat{y} = \Delta \hat{y} +  \bar{y}_{\text{train}}
\end{align}

\subsection{Graph construction}
\label{sec:graph_construction}

% 1. Construccion de los grafos (features visuales, positional encoding (concatenar o sumar), conectividad (k-hop vecinity), bidireccionales y sin pesos en los edges)

The process of building the graphs is shown in Fig.\ref{fig:model_overview}.B and aims to follow the spatial connectivity of the WSI. Therefore, for a patch of interest $X$, we first select the $k$ neighbors within an $m-$hop vicinity of $X$. For example, in Fig.\ref{fig:model_overview}B $m=1$ and $k=6$ because of the hexagonal coordinate geometry. We join the patch and its neighbors in a single set $P = \{X, Z\} \in \mathbb{R}^{[k+1, H, W, 3]}$ and compute the visual embedding matrix $M_i \in \mathbb{R}^{[d_{\text{emb}},k+1]}$ using our frozen image encoder $I(\cdot)$. Additionally, to enrich the spatial information beyond the topology of our graphs, we calculate a positional embedding $E_{\text{pos}} \in \mathbb{R}^{[d_{\text{emb}},1]}$ for each patch in $P$ using the 2D transformer-like positional encoder from \cite{2DEncoder}. The inputs of that encoder are the relative coordinates of each neighbor w.r.t. the center patch. This computation gives us a positional matrix $M_p\in \mathbb{R}^{[d_{\text{emb}},k+1]}$ that is added with $M_i$ to give the final graph features $M$. Summarizing, we define graphs as: 
\begin{align}
    G(X) &= \mathcal{G}(P, E, M)\\
    M&=M_i+M_p
\end{align}

Where $E$ is a binary and undirected set of edges defined by dataset geometry.

\subsection{Spatial Learning Module}

Once a graph $\mathcal{G}(X)$ is fed to the spatial learning module, it is passed through a series of $h$ Graph Convolutional Operators ($\text{GNN}_i(\cdot)$) with a sequence $C=\{d_{\text{emb}}, c_1, c_2, \dots, c_{h-1}, n_g\}$ of hidden channels following the recursive expression:
\begin{align}
    g_0 &= \mathcal{G}(X)\\
    g_{i+1} &= \sigma\left(\text{GNN}_i(g_i)\right)\\
    \hat{s} &= \text{Pooling}(g_h)
\end{align}

Where $g_i$ is the representation of $\mathcal{G}(X)$ at layer $i \in \{0,1,2,\dots,h\}$, $\sigma(\cdot)$ is an activation function, and the $\text{Pooling}(\cdot)$ operator represents a global graph pooling operator. The correction vector $\hat{s}$ represents the contribution of local spatial information to the final prediction.

% Arquitectura del spatial learning
% 1. Posibles capas de preprocesamiento con MLPs
% 2. arquitecturas posibles de las capas de grafos
% 3. Global Pooling (SAGPooling)
% 4. Prediction head with MLP
    
%----------------------------------------------------------------------------------------------------------
\section{Experiments}

\subsection{Datasets}
We evaluate our performance in two breast cancer datasets produced with different technologies: (1) the 10x Genomics breast cancer spatial transcriptomic dataset [\href{https://www.10xgenomics.com/resources/datasets/human-breast-cancer-block-a-section-1-1-standard-1-1-0}{Section 1}, \href{https://www.10xgenomics.com/resources/datasets/human-breast-cancer-block-a-section-2-1-standard-1-1-0}{Section 2}] (referred to as \textit{Visium} because of the experimental protocol), and (2) the human breast cancer in situ capturing transcriptomics dataset \cite{STNet_dataset1, STNet_dataset2} (referred to as \textit{STNet dataset} because of the first deep learning method that used this data). The Visium dataset contains two slide images from a breast tissue sample with invasive ductal carcinoma from one patient, each with $3798$ and $3987$ spots of $\approx55\mu$m detected under the tissue. On the other hand, STNet dataset consists of 68 slide images of H\&E-stained tissue from 23 patients with breast cancer and their corresponding spatial transcriptomics data. Specifically, the number of spots of size $\approx 150\mu$m varies between $256$ and $712$ in each replication, so the complete dataset contains $30,612$ gene expression data points with their respective spatially associated image patch. For both datasets, we take reshaped patches to a $[224, 224, 3]$ dimension as input for \methodname.

\subsection{Benchmark}

To design a robust benchmark, we focus on three main characteristics: (1) a bioinformatic pipeline on par with current best practices in transcriptomic analysis, (2) a pepper noise filter to improve data quality and allow better model training, and (3) a selection strategy to ensure that all genes have spatial patterns. 
 
In terms of the processing pipeline, we first filter out both genes and samples with total counts outside a defined range (See Supplementary Table 1 for detailed values in each dataset). Then, we discard genes based on their sparsity. Here, we ensure that the remaining variables are expressed in at least $\varepsilon_T$ percent of the total dataset and $\varepsilon_{WSI}$ percent of each WSI. Following the filtering, we perform TPM \cite{tpm} gene normalization and a $\log_2(x+1)$ transformation. 

\begin{figure}
    \centering
    \includegraphics[width=0.47\textwidth]{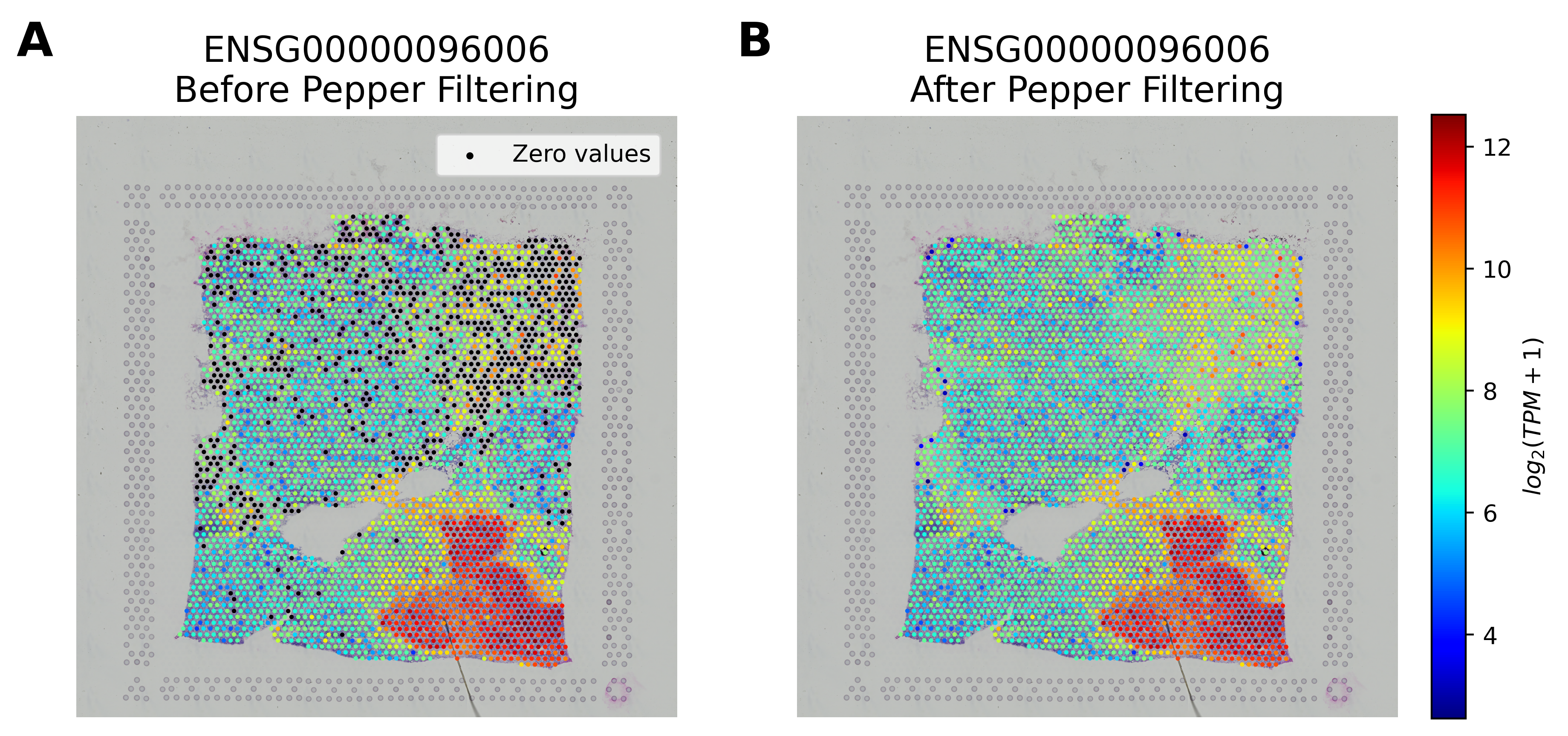}
    \caption{Example of the pepper denoising for a specific gene map in the Visium dataset.}
    \label{fig:pepper_noise}
\end{figure}

To address pepper noise, we applied a modified version of the adaptive median filter \cite{median_filter}. Shortly, for each zero value in a gene map, we replace it with the median of a growing circular region around the interest patch up to the $7^{th}$ unique radial distance. If no value is obtained at the end of this process, we assign the median of nonzero entries of the WSI. The results of this procedure can be appreciated for a particularly noisy gene map in Figure \ref{fig:pepper_noise}. It is worth noting that the percentage of imputed values is $5.3\%$ and $26.0\%$ for the Visium and STNet datasets, respectively, as we have already filtered genes based on their sparsity ($\varepsilon_T, \varepsilon_{WSI}$).

Once the bioinformatic pipeline and the denoising procedure are complete, we select the final prediction variables with the help of Moran's I \cite{moran}. This statistic is a spatial autocorrelation measure and can detect if a given gene has a pattern over spatial graphs. The closer its value to one, the more autocorrelated the variable is. For our benchmark, we compute Moran's I for every gene and WSI and average across the slide dimension. We select the top $n_g=256$ genes with the highest general Moran's I value as our final prediction variables (See supplementary Figures 4-7). Finally, if batch effects are observed in UMAP \cite{umap} embeddings (Supplementary Figures 1-3) of the data (only seen in the STNet dataset), they are corrected with ComBat \cite{combat}.

Summarizing, the processed Visium and STNet datasets have a total of 7,777 and 29,820 samples, respectively, along with a set of 256 prediction genes. As the Visium dataset only contains two WSIs, we use one for training (3795 samples) and the other one as the validation/test set (3982 samples). For the STNet dataset, from the 23 patients, we randomly choose 15 for training (20,734 samples), 4 for validation (3,397 samples), and 4 for testing (5,689 samples).

\subsection{Evaluation Metrics}

We use three standard metrics in multivariate regression problems: global standard errors (MSE, MAE), Pearson Correlation Coefficients (PCC-Gene, PCC-Patch), and linear regression determination coefficients (R2-Gene, R2-Patch). Both PCC and R2 have gene and patch variants since they address two aspects of the problem. The gene type metrics aim to quantify how good expression maps are in general, while the patch type metrics evaluate how good multiple gene predictions are for a specific patch. For instance, to compute PCC-Gene, we obtain PCC values for each one of the $n_g$ gene maps and then average over the gene dimension. Conversely, computing PCC-Patch involves calculating PCC values for each patch and the average over that dimension. Importantly, the imputed values are ignored for metric computation, and consequently, performance measurements are based exclusively on real data.
% The errors are expected to take values close to zero, whereas PCC and R2 should be close to 1.0.

\subsection{State-of-the-art Methods}

We compare \methodname~to four of the most popular methods in this task, including three local options (STNet, EGN, EGGN), as well as one global method (HisToGene). For a fair comparison, we choose the best performance between 50 different training protocols. If the method allows batch size as a hyperparameter (STNet, EGN), we test combinations with an empirical Bayes approach by selecting learning rates in the logarithmic range $[10^{-2},10^{-6}]$ and batch sizes from the list $[32, 64, 128, 256, 320]$. If the method only accepts the learning rate, we perform a logarithmic grid search within the range $[10^{-2},10^{-6}]$. Both the best epoch during training and the best model of the sweep are selected based on the validation MSE. The only exception to this protocol (due to computational cost) is the STNet method in the STNet dataset, for which we report the best between the original hyperparameters and the best Visium hyperparameters.

\subsection{Architecture Optimization}

We extensively experiment with our spatial module, aiming to select the most effective architecture to integrate local information. For this purpose, we: (1) optionally introduce pre-processing and post-processing stages via multilayer perceptrons of varying sizes, (2) allow the positional encoding to be added or concatenated during the graph construction, (3) change the number of hops $m$ from one to three, (4) try six different convolutional operators, and (5) vary the hidden dimensions $h$ of our graph convolutional network going from one to four layers. Furthermore, we train all architecture variations with 12 different settings of learning rate and batch size. For a detailed explanation of every tunned hyperparameter, we refer the reader to the Supplementary Material (Sec. 2).

\subsection{Implementation Details}

After comprehensive experimentation (Supplementary Material Sec. 3), we choose ViT-B-16 \cite{ViT} as our image encoder. We use ELU as our activation function, and SAGPooling \cite{sagpooling} as our pooling function. We train in the denoised version of the dataset but only use real data for metric computation during inference. We implement \methodname~using Pytorch \cite{pytorch} and Pytorch geometric \cite{pytorch_geometric} for graph operators. All experiments run on a single NVIDIA Quadro RTX 8000 GPU.

%----------------------------------------------------------------------------------------------------------
\section{Results}

\begin{figure*}[ht]
    \centering
    \includegraphics[width=0.99\textwidth]{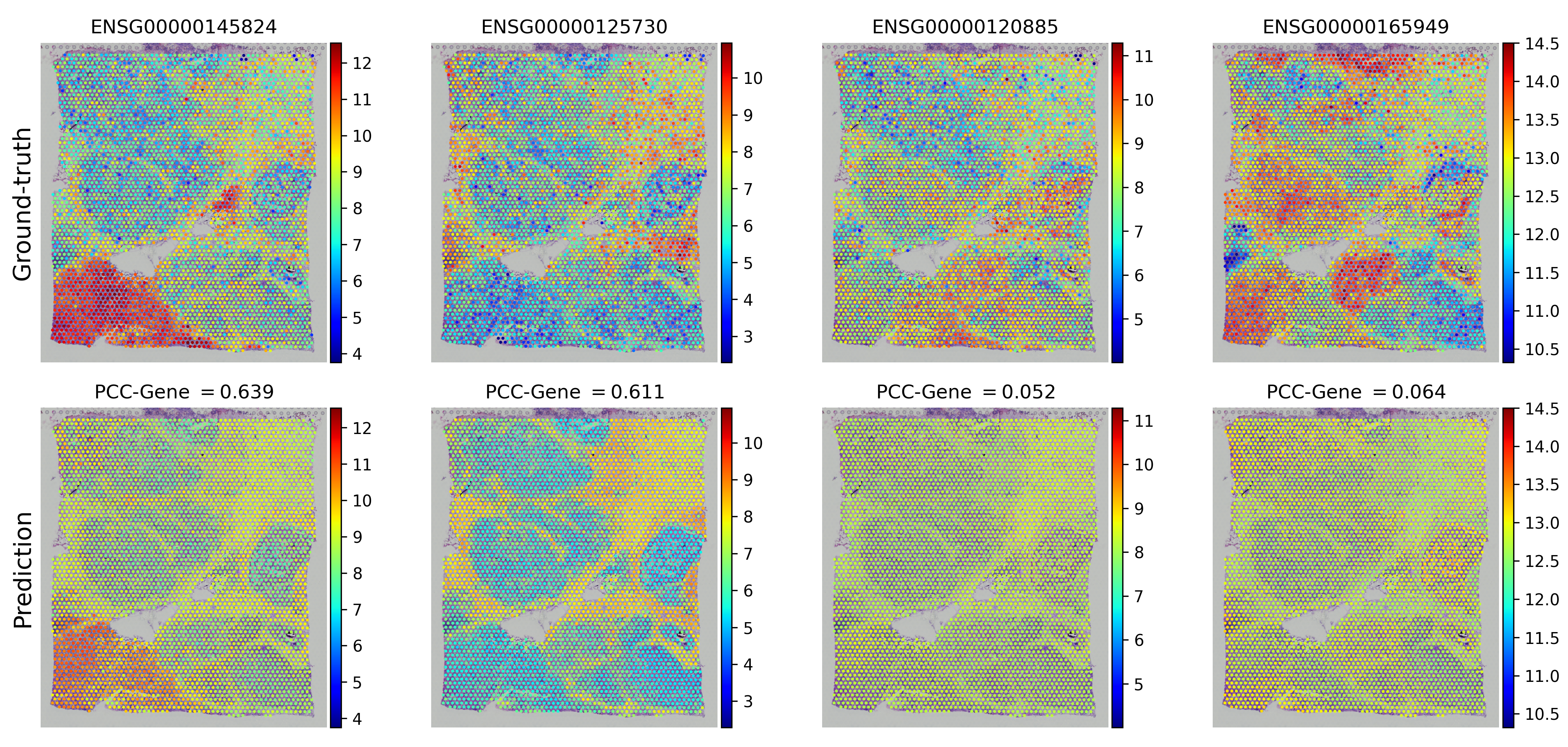}
    \caption{Visualization of the two genes with the highest (left) and lowest (right) Pearson Correlation Coefficient. At the top is the Ground-Truth of the expression and at the bottom is the qualitative prediction of our method with its respective PCC.}
    \label{fig:qualitative_results}
\end{figure*}
% El gen que mejor se comporta se predice bien aun cuando su expresion no solo esta determinada por el color, que es el caso de el segundo mejor gen. Por lo tanto la red parece estar aprendiendo de la morfologia especifia observada
% Se nota que los outputs de los genes buenos no tienen el mismo rango dinamico de las anotaciones
% La salida de nuestro metodo es en general mucho mas suave que su anotacion (the output of aur method is oversmothed with respect to the groundturth). Esto es bueno porque implica consistencia en la prediction (entradas parecidas -> salidas parecidas) pero puede ser malo si se busca detectar una desviacion especifica de un gen con alta resolucion espacial.
% A los genes a los que les fue mal se observa tambien que incluso en la anotacion el patron espacial no es muy definido
% El caso de fallo observado es que en general los genes malos se quedan en la media de la expresion y generan predicciones casi constantes que no funcionan bien

\begin{figure}
    \centering
    \includegraphics[width=0.47\textwidth]{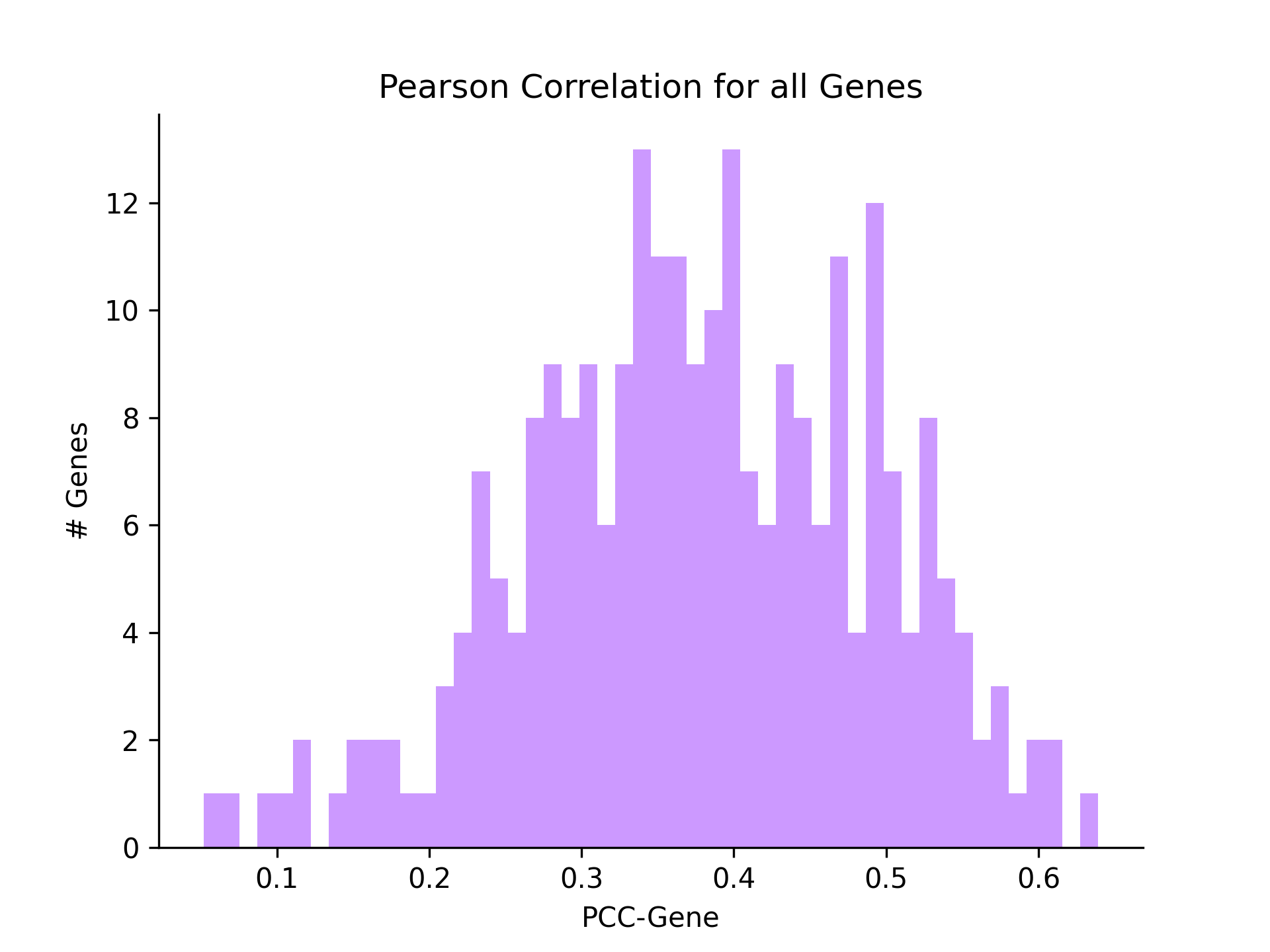}
    \caption{Histogram of the Pearson correlation between the ground-truth and the prediction of each gene. The X-axis displays the values of the Pearson correlation coefficient, while the Y-axis shows the number of genes that have that particular correlation.}
    \label{fig:pcc_histogram}
\end{figure}
% Distribucion aproximadamente normal por inspeccion visual
% QUe no hay outliers importantes 
% Todos los genes tienen una correlacion mayor a 1
% El rango es aprox 0.1, 0.66

Table \ref{tab:hyperparameters} presents the final hyperparameter configurations of \methodname~for the Visium and the STNet datasets. %The most interesting observation is that the amount of context needed to obtain optimal results varies between datasets (1 hop Vs 3 hops). This result is expected since the spatial resolution of the STNet dataset is lower than that of the Visium dataset, meaning that with a single hop our model can obtain much more information.   

\begin{table}[h]
\resizebox{0.48\textwidth}{!}{
\begin{tabular}{@{}ccc@{}}
\toprule
\textbf{Hyperparameters} & \textbf{STNet Dataset} & \textbf{Visium} \\ \midrule
Number of hops & 1 & 3 \\
Embeddings aggregation & Sum & Concat \\
Graph operator & GraphConv\cite{GraphConv} & GCNConv\cite{GCNConv} \\
Preprocessing Stage & - & $d_\text{emb}$, 512 \\
Graph hidden channels & $d_\text{emb}$, 256 & 512, 256, 128 \\
Postprocessing Stage & - & 128, 256 \\
Learning rate & $10^{-4}$ & $10^{-5}$ \\
Batch size & 256 & 256 \\ \bottomrule
\end{tabular}}
\vspace{0.1cm}
\caption{Hyperparameters with the best performance for both datasets.}
\label{tab:hyperparameters}
\end{table}

\subsection{Main Results}

% \usepackage{booktabs}
% \usepackage{multirow}
\begin{table}[h!]
\centering
\resizebox{0.48\textwidth}{!}{
\begin{tabular}{@{}cccccccc@{}}
\toprule
 &  & \multicolumn{3}{c}{\textbf{Local}} & \textbf{Global} & \multicolumn{2}{c}{\textbf{Hybrid}} \\ \midrule
 & \multicolumn{1}{c|}{\textbf{Method}} & \textbf{STNet} \cite{STNet} & \textbf{EGN} \cite{EGN} & \multicolumn{1}{c|}{\textbf{EGGN} \cite{EGGN}} & \multicolumn{1}{c|}{\textbf{HisToGene} \cite{HisToGene}} & \textbf{SEPAL} & \textbf{SEPAL*} \\ \midrule
\multirow{6}{*}{\rotatebox{90}{Visium}}& \multicolumn{1}{c|}{MAE} & 0.654 & 0.659 & \multicolumn{1}{c|}{0.648} & \multicolumn{1}{c|}{0.665} & \textbf{0.630} & \underline{0.636} \\
 & \multicolumn{1}{c|}{MSE} & 0.762 & 0.772 & \multicolumn{1}{c|}{0.741} & \multicolumn{1}{c|}{0.784} & \textbf{0.708} & \underline{0.717} \\
 & \multicolumn{1}{c|}{PCC-Gene} & 0.300 & 0.314 & \multicolumn{1}{c|}{0.310} & \multicolumn{1}{c|}{0.199} & \textbf{0.383} & \underline{0.353} \\
 & \multicolumn{1}{c|}{R2-Gene} & 0.053 & 0.038 & \multicolumn{1}{c|}{0.068} & \multicolumn{1}{c|}{0.024} & \textbf{0.106} & \underline{0.091} \\
 & \multicolumn{1}{c|}{PCC-Patch} & 0.924 & 0.922 & \multicolumn{1}{c|}{0.925} & \multicolumn{1}{c|}{0.921} & \textbf{0.928} & \underline{0.927} \\
 & \multicolumn{1}{c|}{R2-Patch} & 0.843 & 0.841 & \multicolumn{1}{c|}{0.845} & \multicolumn{1}{c|}{0.839} & \textbf{0.853} & \underline{0.851} \\ \midrule
\multirow{6}{*}{\rotatebox{90}{STNet dataset}} & \multicolumn{1}{c|}{MAE} & 0.560 & \underline{0.520} & \multicolumn{1}{c|}{0.545} & \multicolumn{1}{c|}{0.529} & \textbf{0.519} & 0.527 \\
 & \multicolumn{1}{c|}{MSE} & 0.537 & \underline{0.480} & \multicolumn{1}{c|}{0.518} & \multicolumn{1}{c|}{0.493} & \textbf{0.478} & 0.489 \\
 & \multicolumn{1}{c|}{PCC-Gene} & \underline{0.030} & \textbf{0.064} & \multicolumn{1}{c|}{0.014} & \multicolumn{1}{c|}{-0.007} & -0.004 & 0.002 \\
 & \multicolumn{1}{c|}{R2-Gene} & -0.165 & \underline{-0.037} & \multicolumn{1}{c|}{-0.129} & \multicolumn{1}{c|}{-0.066} & \textbf{-0.028} & -0.052 \\
 & \multicolumn{1}{c|}{PCC-Patch} & \underline{0.910} & \textbf{0.911} & \multicolumn{1}{c|}{0.907} & \multicolumn{1}{c|}{\textbf{0.911}} & \textbf{0.911} & \textbf{0.911} \\
 & \multicolumn{1}{c|}{R2-Patch} & 0.779 & \underline{0.806} & \multicolumn{1}{c|}{0.792} & \multicolumn{1}{c|}{0.799} & \textbf{0.809} & 0.802 \\ \bottomrule
\end{tabular}}
\vspace{0.1cm}
\caption{Quantitative comparison with state-of-the-art methods on Visium and STNet datasets. The best performance is written in \textbf{bold}, and the second best result is \underline{underlined} for each metric. *: SEPAL architecture with optimal parameters from the other dataset}
\label{tab:main_results}
\end{table}

Table~\ref{tab:main_results} depicts the  performance of local and global state-of-the-art methods against \methodname~on the Visium and STNet datasets. 
Our method consistently outperforms these methods on all but one evaluation metric. 
% when working on the Visium dataset, and so is the result for the STNet dataset except for the MAE, where \methodname~is 0.002 points behind the best-performing model. 
% Our method successfully decreases the MSE to 0.029 points lower than EGGN on Visium and  0.013 points lower than the STNet model on the STNet dataset. 
In particular, we attend primarily to the standard error metrics and find that \methodname~presents significant improvements on the Visium dataset and performance on par with state-of-the-art for the STNet dataset. Likewise, the R2 metric calculated on the genes increases for both datasets when using \methodname. % For this metric, however, the major increment is on the Visium dataset, noting that our model has an R2-Gene of 0.019 points higher than the state-of-the-art. 

Furthermore, we find that calculating the PCC and the R2 metrics in a gene-wise fashion results in a remarkably poorer performance compared to the patch-wise evaluation. This means that predicting the distribution of the expression of a single gene in a WSI is a significantly more difficult task than aiming to obtain the expression of all the genes in one single spot. Nevertheless, despite this different trend for gene or patch-focused evaluations, our method consistently achieves the best results.

% As an additional observation, for the STNet dataset, the STNet model shows consistent results being the second best model in almost every evaluation metric, while on the Visium dataset there is not one single method that reports this behaviour. Besides, it is worth noting that for both datasets the performance of all the methods on all the metrics remains within a close range. This shows that there is no critical difference in the performance of the global and the local methods. Nonetheless, 
HisToGene has poorer performance on Visium than on STNet, and overall it shows the worst results on the Visium dataset. These differences within the results of HisToGene support the observation that data scarcity of small datasets like Visium leads to deficient results in global methods. Conversely, we demonstrate that our method is able to retrieve important information from the input despite the difference in the data acquisition technologies and number of samples since it achieves high performance on both datasets.

Finally, Fig.\ref{fig:pcc_histogram} shows a histogram of the PCC between the ground-truth and the predictions of each gene on Visium. None of the genes has a negative correlation, and the lowest PCC is 0.052 and goes as far as 0.639. Overall, our model has a satisfactory performance for the evaluation of the genes selected. We observe that the PCC has an approximately normal distribution, with no evident outliers.

As a sidenote observation, we also validated our data imputation protocol by obtaining the main results for the Visium dataset when training in noisy data. The metrics show a consistent drop in performance disregarding the method (Sec. 4 Supplementary Material), which supports the need for our denoising approach and sets a best practice for future works.

\subsection{Control Experiments}

\begin{table}[]
\centering
\resizebox{0.45\textwidth}{!}{
\begin{tabular}{@{}lccccc@{}}
\toprule
\multicolumn{1}{c}{\textbf{Method}} & \textbf{ViT} & \textbf{ViT$+\Delta$} & \textbf{ViT$+\Delta+$S7} & \textbf{\methodname} \\ \midrule
MAE & 0.655 & \underline{0.638} & 0.648 & \textbf{0.630} \\
MSE & 0.760 & \underline{0.725} & 0.737 & \textbf{0.708} \\
PCC-Gene & 0.282 & \underline{0.347} & 0.339 & \textbf{0.383} \\
R2-Gene & 0.053 & \underline{0.086} & 0.065 & \textbf{0.106} \\
PCC-Patch & 0.924 & \underline{0.927} & 0.925 & \textbf{0.928} \\
R2-Patch & 0.843 & \underline{0.849} & 0.847 & \textbf{0.853} \\ \bottomrule
\end{tabular}}
\vspace{0.2cm}
\caption{Control experiments on the Visium validation/test set. $\Delta$: predicting differences with respect to the mean expression $\bar{y}_{\text{train}}$. S7: input patch is 7 times bigger than the original one.}
\label{tab:ablation_results}
\end{table}

Table \ref{tab:ablation_results} shows the results for the ablation experiments. Comparing the results between predicting the absolute expression (ViT) and predicting the expression variations of the genes (ViT$+\Delta$), we notice that the latter option has a better performance in every metric. For instance, when predicting delta variations, the MSE is 0.035 points below that of the absolute expression prediction. The PCC-Gene also increased 0.065 points with our problem formulation. These results reflect the suitability of the paradigm shift that we propose by learning the difference between $y$ and $\bar{y}_{\text{train}}$ instead of directly predicting $y$.

We evaluate the benefit of using a larger neighborhood to determine how raw spatial information affects gene prediction (ViT+$\Delta$ against ViT+$\Delta$+S7).
Table \ref{tab:ablation_results} compares the behavior of the exact same image encoder while solely altering the scale of the patches (with seven times more visual context). For all metrics, keeping a scale of 1.0 remains the best option among the ViT architectures tested. Our findings suggest that increasing the visual coverage of an image encoder does not yield significant improvements in gene prediction. 

In addition, the results from \methodname~show an improvement over all metrics, with respect to ViT+$\Delta$+S7. Notably, both \methodname~and ViT+$\Delta$+S7 have access to the same visual context in the WSI and are differentiated only by how spatial information is represented.
% We see that \methodname~has a MAE, a MSE and a R2-Gene that are 0.005, 0.013 and 0.015 points better than the second best values, respectively. 
This compelling outcome underscores the importance of incorporating spatial features in the description of each patch and constructing graphs to glean highly relevant information for accurate expression prediction.
The performance of \methodname~shows that predictions from the spatial module do further improve the preliminary predictions obtained during the local learning stage. Our results validate the efficacy of our novel approach, emphasizing the value of spatial interactions in gene expression prediction.

% As a sidenote observation, when predicting on the STNet dataset, ViT+$\Delta$ converges in earlier stages of the training process than ViT, and the training of the model is notably more stable than when predicting absolute gene expression.

\begin{comment}
Adicionar el delta en STNet dataset aunque no mejoro mucho las metricas si mejoro mucho la rapidez y estabilidad del entranamiento (todas las 20 runs estuvieron en el orden de magnitude del ganador approx 0.6 mientras queue sin el delta el error inicial estuvo en el orden de magnitude de 60 para al menos la mitad de los runs)
\end{comment}

\subsection{Qualitative Results}
Figure.\ref{fig:qualitative_results} shows the heatmaps for the real and the predicted expression distribution of the genes with the best and worst performances. Focusing on the genes with the highest PCC, we see that for the second best gene, the expressions both on the ground-truth and on the prediction are highly associated with the tissue color. Note that the regions with darker tissue obtain lower expression predictions, and regions with lighter tissue obtain higher expression predictions. These results suggest that our model might be basing estimates solely on the color of the patches rather than looking for specific morphology patterns. Nevertheless, for the best gene, the predicted expressions are not uniformly the same for all dark or light tissue sections, conveying that our model does not rely only on the tone of images and is actually learning from the spatial context of the patches and tissue morphology.

The predicted expressions show a lower intensity than the ground-truth for both genes, indicating that the dynamic range of \methodname~predictions may not match that of the real expression levels. Notably, for the two genes with the highest PCC, the output of our method appears over-smoothed compared to the ground-truth. 
An evident distinction arises when comparing the real expression, which exhibits adjacent spots with drastically different expression levels, to the predictions, where no regions display sudden changes in expression tendencies. While our model's consistent predictions showcase its strength, this attribute may also be considered a drawback when seeking to detect gene expression deviations with high spatial resolution. 
% We can clearly notice that the real expression has adjacent spots whose expression levels are drastically different. In contrast, in the predictions, there are no regions where the expression tendency shows sudden changes.
% Our model presents consistent predictions but can also be regarded as a drawback if we want to detect gene expression deviations with a high spatial resolution.

Regarding the hardest genes, Fig.\ref{fig:qualitative_results} shows that the predictions tend to correspond to the mean expression value of each gene and are practically constant throughout the entire WSI. For these cases, SEPAL fails to capture the spatial dependencies, even though clear patterns are present in the ground-truths. We hypothesize that these shortcomings are due to the joint prediction of 256 genes.
% in their ground-truth that these are cases where the expression does not have a particularly clear spatial pattern, even if these genes surpassed our Moran's I Statistic threshold. Consequently, our model's poor performance on these genes can be attributed to the absence of a major spatial pattern to predict. Additionally, we see that for bad-performing genes, the predictions tend to correspond to the mean expression value of each gene and are practically constant throughout the entire WSI. 

%----------------------------------------------------------------------------------------------------------
\section{Conclusions}
\begin{comment}
    \1 In this work, we develop a novel framework to approach the gene expression prediction task by integrating local spatial context, and exploiting inductive biases inherent to the biological nature of the problem.
    \1 Our proposed \methodname, consistently outperforms state-of-the-art models and closes the gap between completely global and completely local analysis. Furthermore, it is capable of recognizing patterns in histological data that go beyond simple color intensities, as is expected biologically. 
    \1 Consequently, our approach represents a significant step forward in spatial expression prediction, enhancing the applicability of deep learning methods in the context of disease analysis and precision medicine.
\end{comment}

In this work, we develop a novel framework to approach the spatial gene expression prediction task by integrating local context and exploiting inductive biases inherent to the biological nature of the problem. Our proposed \methodname~consistently outperforms state-of-the-art models and closes the gap between completely global and completely local analysis. Furthermore, aligning with biological expectations, it is capable of recognizing patterns in histological data that go beyond simple color intensities. Consequently, our approach represents a significant step forward in spatial expression prediction, enhancing the applicability of deep learning methods in the context of disease analysis and precision medicine.

% \angela{Idea para conclusión}
% we achieve a more robust and context-aware prediction process, considering the physiological significance of relative expression changes within the gene expression range. This not only aligns with biological expectations but also enables us to glean valuable insights into the dynamic behavior of genes within tissues. 
% Consequently, our approach represents a significant step forward in spatial expression prediction, enhancing the interpretability and applicability of our results in the context of disease analysis and precision medicine.

\section{Acknowledgements}

Gabriel Mejia acknowledges the support of a UniAndes-DeepMind Scholarship 2022. This work was supported by Azure sponsorship credits granted by Microsoft’s AI for Good Research Lab. We thank Andres Hernandez for his valuable help during the conception of the project.

{\small
\bibliographystyle{ieee_fullname}
\bibliography{egbib}
}

\end{document}